\documentclass[cameraready]{Interspeech}

\title{Preference-ASR: A Preference-Aware Test Set for Benchmarking ASR in the Era of Speech LLMs}

\author[affiliation={1}]{Nithin Rao}{Koluguri}
\author[affiliation={1}]{Sasha}{Meister}
\author[affiliation={1}]{Nikolay}{Karpov}
\author[affiliation={1}]{Piotr}{Zelasko}
\author[affiliation={1}]{Desh}{Raj}
\author[affiliation={1}]{Jagadeesh}{Balam}
\author[affiliation={1}]{Boris}{Ginsburg}

\address{
    $^1$ NVIDIA, USA
}
\email{nkoluguri@nvidia.com}


\usepackage{comment}
\usepackage{multirow}
\usepackage{booktabs}
\usepackage{tikz}
\usetikzlibrary{positioning, arrows.meta, fit, backgrounds}
\usepackage{pifont}
\usepackage{todonotes}

\usepackage{xcolor}

\usepackage{caption}
\usepackage{eurosym}
\definecolor{normbg}{HTML}{E3F2FD}
\definecolor{normbdr}{HTML}{1976D2}
\definecolor{entbg}{HTML}{F3E5F5}
\definecolor{entbdr}{HTML}{8E24AA}
\definecolor{disfbg}{HTML}{FFF3E0}
\definecolor{disfbdr}{HTML}{E67E22}
\definecolor{casebg}{HTML}{E8F5E9}
\definecolor{casebdr}{HTML}{388E3C}

\begin{document}

\maketitle

\begin{abstract}
    Popular ASR test sets adopt inconsistent conventions for numbers, disfluencies, entities, and casing, while standard normalizers erase the format distinctions users care about. Current benchmarks therefore cannot measure whether a model follows user preferences for output style. We introduce Preference-ASR, a test set evaluating ASR systems on their ability to follow natural-language preference instructions across four categories: normalization, entities, disfluencies, and case. Built from seven open-source corpora via a two-stage LLM-assisted pipeline with human verification, it is evaluated with a preference-aware normalizer that selectively skips steps matching the active instruction. Benchmarking four models shows rankings shift across preference types, exposing quality differences traditional evaluation obscures. We publicly release the dataset\footnote{\url{https://github.com/nithinraok/preference-asr-bench}}.
\end{abstract}

\section{Introduction}

Automatic speech recognition (ASR) has evolved rapidly, from hidden Markov models through end-to-end CTC and RNN-Transducer architectures to the recent class of Speech-augmented Large Language Models (SpeechLLMs) such as SALMONN~\cite{tang2024salmonn}, Qwen-Audio~\cite{chu2023qwenaudio}, and Qwen2.5-Omni~\cite{xu2025qwen25omni}. Unlike earlier systems, SpeechLLMs can follow natural-language instructions that specify the desired style and format of a transcription. Yet the way we evaluate ASR has not kept pace: we still measure Word Error Rate (WER) against fixed human-annotated references, an approach that increasingly fails to capture what modern models can actually do.

One root cause is that annotation conventions vary widely across popular test sets and are seldom documented in full. Number formatting alone illustrates the problem: Librispeech~\cite{panayotov2015librispeech} writes numbers in spoken form and lowercases everything, SPGISpeech~\cite{oneill2021spgispeech} uses inverse-normalized written form but omits proper casing, and Earnings-22~\cite{delrio2022earnings22} mixes both forms within the same dataset. Disfluency handling is equally inconsistent: some corpora retain every filler word, repetition, and false start, while others such as SPGISpeech strip them entirely. Speaker conventions diverge too: GigaSpeech~\cite{chen2021gigaspeech} contains multi-speaker audio yet provides only one speaker's transcription, and AMI~\cite{ami} mostly ignores background speech without a clear rule for which speaker is the reference.
VoxPopuli~\cite{wang2021voxpopuli} and Common Voice~\cite{ardila2020commonvoice} add further alignment and transcription quality issues. Figure~\ref{fig:motivation} illustrates two such inconsistencies: GigaSpeech writes numbers in spoken form while Earnings-22 uses written form, and AMI preserves filler words while SPGISpeech strips them. The net effect is that a model's ranking on any single benchmark often reflects how well it happens to match that dataset's annotation quirks, rather than its genuine transcription quality.

These inconsistencies matter more now that SpeechLLMs can act on user instructions. In practice, users have specific formatting preferences: a database pipeline may need digits (Inverse Text Normalization, ITN), while a broadcast script calls for words (Text Normalization, TN). A pronunciation coaching app needs every disfluency preserved, whereas a meeting summary should omit them. Users may also want to supply domain-specific entity names (pharmaceutical terms, product names, proper nouns) to improve recognition accuracy, or request a specific casing style. No existing test set can express these preferences. Worse, standard evaluation pipelines apply blanket normalization (lowercasing, number-to-word conversion, symbol removal) that erases precisely the formatting distinctions users care about, treating them as noise rather than as meaningful signal of system capability.

\definecolor{normbg}{HTML}{E3F2FD}
\definecolor{normbdr}{HTML}{1976D2}
\definecolor{disfbg}{HTML}{FFF3E0}
\definecolor{disfbdr}{HTML}{E67E22}
\begin{figure}[t]
    \centering
    \begin{tikzpicture}[
      font=\footnotesize,
      exbox/.style={draw, rounded corners=2pt, semithick, inner sep=3pt},
    ]
    \node[exbox, fill=normbg, draw=normbdr] (norm) {\parbox{0.95\columnwidth}{%
      \textbf{\textcolor{normbdr}{Normalization}}\\[2pt]
      \textbf{GigaSpeech GT:} \ldots a similar problem exists with the {\textbf{\textcolor{normbdr}{seventy percent}}} rule \ldots\\[1pt]
      \textbf{Earnings-22 GT:} In the quarter you come up strong as thought, {\textbf{\textcolor{normbdr}{10\%}}}.\\[1pt]
      \hfill{\scriptsize\textit{\textcolor{normbdr}{GigaSpeech uses spoken form; Earnings-22 uses written form}}}}};

    \node[exbox, fill=disfbg, draw=disfbdr, below=2pt of norm] (disf) {\parbox{0.95\columnwidth}{%
      \textbf{\textcolor{disfbdr}{Disfluencies}}\\[2pt]
      \textbf{AMI GT:} {\textbf{\textcolor{disfbdr}{Um}}} and my findings from this is that, {\textbf{\textcolor{disfbdr}{you know}}}, small is beautiful.\\[1pt]
      \textbf{SPGISpeech GT:} And my findings from this is that, small is beautiful.\\[1pt]
      \hfill{\scriptsize\textit{\textcolor{disfbdr}{AMI keeps filler words; SPGISpeech strips them}}}}};

    \end{tikzpicture}
    \vspace{-4pt}
    \caption{Existing ASR references exhibit inconsistent conventions across datasets: GigaSpeech writes numbers in spoken form while Earnings-22 uses written form; AMI preserves filler words while SPGISpeech strips them. \textbf{Preference-ASR} resolves this by conditioning each reference on an explicit user instruction.}
    \label{fig:motivation}
\end{figure}

\begin{figure*}[!t]
\centering
\definecolor{graybg}{HTML}{F5F5F5}
\definecolor{graybdr}{HTML}{888888}
\definecolor{orangebg}{HTML}{FFF3E0}
\definecolor{orangebdr}{HTML}{E67E22}
\definecolor{bluebg}{HTML}{E3F2FD}
\definecolor{bluebdr}{HTML}{1976D2}
\definecolor{purplebg}{HTML}{F3E5F5}
\definecolor{purplebdr}{HTML}{8E24AA}
\definecolor{greenbg}{HTML}{E8F5E9}
\definecolor{greenbdr}{HTML}{388E3C}
\resizebox{\textwidth}{!}{%
\begin{tikzpicture}[
  font=\small,
  node distance=0.4cm,
  box/.style={draw, rounded corners=3pt, thick, inner sep=5pt,
              text width=#1, align=center, minimum height=1.0cm},
  box/.default=2.0cm,
  stage/.style={draw, rounded corners=6pt, thick, dashed, inner sep=8pt},
  arw/.style={-{Stealth[length=5pt]}, thick},
]

\node[box=2.1cm, fill=graybg, draw=graybdr] (input) {%
  \textbf{Audio}\\[1pt]$+$\\[1pt]\textbf{Ground Truth}\\[-1pt]
  {\scriptsize(7 corpora,}\\[-2pt]{\scriptsize 3{,}545 samples)}};

\node[box=2.1cm, fill=orangebg, draw=orangebdr, right=0.5cm of input] (mv1) {%
  {\small\ding{51}}\\[1pt]\textbf{Manual}\\[1pt]\textbf{Verification}\\[-1pt]
  {\scriptsize GT correction}};

\node[box=2.4cm, fill=bluebg, draw=bluebdr, right=0.7cm of mv1] (s1) {%
  \textbf{Preference}\\[1pt]\textbf{Classification}\\[2pt]
  {\scriptsize LLM-assisted}\\[-1pt]
  {\scriptsize Per.\ category}};

\node[box=2.4cm, fill=purplebg, draw=purplebdr, right=0.7cm of s1] (s2) {%
  \textbf{Instruction~+}\\[1pt]\textbf{Reference Gen.}\\[2pt]
  {\scriptsize LLM-assisted}\\[-1pt]
  {\scriptsize Per Preference}};

\node[box=2.1cm, fill=orangebg, draw=orangebdr, right=0.7cm of s2] (mv2) {%
  {\small\ding{51}}\\[1pt]\textbf{Manual}\\[1pt]\textbf{Verification}\\[-1pt]
  {\scriptsize Human review}};

\node[box=2.2cm, fill=greenbg, draw=greenbdr, right=0.5cm of mv2] (output) {%
  \textbf{Preference-ASR}\\[1pt]\textbf{Test Set}\\[-1pt]
  {\scriptsize 3{,}210 triples}\\[-2pt]
  {\scriptsize (audio, instr., ref.)}};

\draw[arw] (input) -- (mv1);
\draw[arw] (mv1)   -- (s1);
\draw[arw] (s1)    -- (s2);
\draw[arw] (s2)    -- (mv2);
\draw[arw] (mv2)   -- (output);

\node[below=0.35cm of s1, font=\small\bfseries, text=bluebdr] {Stage~1};
\node[below=0.35cm of s2, font=\small\bfseries, text=purplebdr] {Stage~2};

\begin{scope}[on background layer]
  \node[stage, draw=bluebdr!60, fill=bluebg!30,
        fit=(s1), inner sep=8pt, yshift=-4pt] {};
  \node[stage, draw=purplebdr!60, fill=purplebg!30,
        fit=(s2), inner sep=8pt, yshift=-4pt] {};
\end{scope}

\end{tikzpicture}%
}
\caption{Two-stage construction pipeline for the Preference-ASR dataset.
Audio samples and ground truth from seven corpora are first \emph{manually verified}.
\textbf{Stage~1} classifies each sample into preference categories (normalization, entities, disfluencies, case) using LLM.
\textbf{Stage~2} generates task-specific instructions and reference texts.
A final round of \emph{manual verification} produces the released test set with  preference-annotated triples.}
\label{fig:pipeline}
\end{figure*}
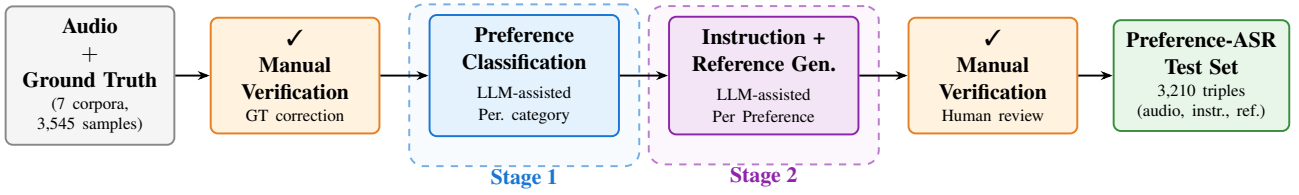

Prior work has touched on parts of this problem. Rich transcription~\cite{meng2021richasr} explored end-to-end formatted ASR output but did not tie evaluation to explicit user instructions. Multi-modal audio benchmarks such as MMAU~\cite{sakshi2024mmau} and Speech-IFEval~\cite{lu2025speechifeval} test understanding and reasoning over audio but do not target the transcription-level preferences central to ASR. Lai et al.~\cite{lai2023instruction} demonstrated instruction-following for speech models, though without addressing the cross-dataset inconsistencies we identify here or releasing a public test set for broader benchmarking. None of these efforts provide a test set whose ground truth is conditioned on a human-specified formatting instruction for ASR.

We introduce \textbf{Preference-ASR}, a test set that evaluates ASR systems on their ability to follow human preference instructions. It draws from seven widely used open-source corpora, combining their strengths while explicitly controlling for the inconsistencies described above. Preference-ASR is not meant to replace existing benchmarks such as the Open ASR Leaderboard~\cite{openasrleaderboard}; rather, it complements them by evaluating a capability that standard test sets cannot measure: whether a model follows user instructions for ASR output formatting. Our contributions are as follows:

\begin{enumerate}
    \item A preference-annotated English test set covering four preference categories: normalization, entities, disfluencies, and case.
    \item A two-stage LLM-assisted pipeline for generating the test set, with human verification and correction.
    \item A preference-aware normalizer that selectively skips normalization steps matching the active preference instruction, enabling fair WER computation across diverse formatting requirements.
    \item We publicly release the test set and evaluation code to support reproducible benchmarking of SpeechLLMs on preference-following.
\end{enumerate}

\section{Preference-ASR Dataset}
\label{sec:dataset}


\subsection{Preference Categories}

We organize preferences into four categories that capture the most common points of friction in real-world transcription workflows. Each category is further divided into sub-categories that can be combined in a single instruction.

\textbf{Normalization.}
Normalization deals with non-standard words whose spoken and written forms differ~\cite{sproat2016rnn}. Choosing between Text Normalization (TN, written to spoken form) and Inverse Text Normalization (ITN, spoken to written form) directly affects downstream tasks. We define three sub-categories:
\begin{itemize}
    \item \textit{Numbers}: Instructions may require digits (ITN, e.g., ``22'') or spoken words (TN, e.g., ``twenty-two'').
    \item \textit{Symbols}: Models may be asked to produce currency symbols and mathematical notation (ITN, e.g., ``\$100'', ``20\%'') or their verbal equivalents (TN, e.g., ``one hundred dollars'', ``twenty percent'').
    \item \textit{Website Links}: Instructions can specify a readable URL (ITN, e.g., ``www.google.com'') or the exact sequence of letters and words spoken (TN, e.g., ``www dot google dot com'').
\end{itemize}
Together, these sub-categories test whether a model can handle dates, currencies, and technical identifiers through a simple prompt rather than hard-coded post-processing rules.

\textbf{Entities.}
Named entities are a persistent source of errors, especially for rare or recently coined names in the tail of the training distribution. Contextual biasing, where a list of likely terms is provided to the model, has been shown to reduce entity word error rate significantly~\cite{pundak2018deep}. Our entity category spans company, product, people, location, organization, event, and drug names. We also include ``false positive'' entities in some instructions: names that are semantically plausible but absent from the audio. This lets us measure confabulation,
a common failure mode where a model favors its textual prior over the acoustic evidence.

\textbf{Disfluencies.}
Spontaneous speech is inherently messy, filled with false starts, repairs, and fillers~\cite{shriberg2005spontaneous}. Whether to keep or remove them depends on the application: a pronunciation coaching tool needs verbatim fidelity, while a meeting summary benefits from cleaning. Sub-categories include:
\begin{itemize}
    \item \textit{Filler Words}: Hesitation markers like ``um,'' ``uh,'' and ``ah.''
    \item \textit{Repetitions}: Consecutive spoken instances of the same word (e.g., ``I I think'').
    \item \textit{False Starts}: Incomplete phrases that are immediately corrected (e.g., ``I was go- going'').
    \item \textit{Wrong Grammar}: Cases where the speaker makes a grammatical error that a model might ``helpfully'' but incorrectly rectify.
\end{itemize}
By framing instructions to either ``keep'' or ``ignore'' these disfluencies, we evaluate whether a model can perform Rich Transcription tasks~\cite{meng2021richasr}.

\textbf{Case.}
Casing and punctuation affect both readability and the accuracy of downstream text analysis~\cite{tilk2016punctuation}. This category covers:
\begin{itemize}
    \item \textit{Lower case}: All letters are lowercased and punctuation is omitted.
    \item \textit{Proper punctuation and capitalization}: Sentences begin with a capital letter, end with appropriate punctuation, and follow standard capitalization rules~\cite{koluguri2024longer}.
\end{itemize}
These categories are mutually exclusive, testing the model's ability to switch its output mode completely based on a formatting directive. Figure~\ref{fig:examples} shows examples from each category.

\begin{figure}[!t]
    \centering
    \definecolor{normbg}{HTML}{E3F2FD}
    \definecolor{normbdr}{HTML}{1976D2}
    \definecolor{entbg}{HTML}{F3E5F5}
    \definecolor{entbdr}{HTML}{8E24AA}
    \definecolor{disfbg}{HTML}{FFF3E0}
    \definecolor{disfbdr}{HTML}{E67E22}
    \definecolor{casebg}{HTML}{E8F5E9}
    \definecolor{casebdr}{HTML}{388E3C}
    \newcommand{\exw}{0.95\columnwidth}%
    \begin{tikzpicture}[
      font=\fontsize{7.7}{5}\selectfont,
      exbox/.style={draw, rounded corners=2pt, semithick, inner sep=3pt},
    ]
    \node[exbox, fill=normbg, draw=normbdr] (norm) {\parbox{\exw}{%
      \textbf{\textcolor{normbdr}{Normalization}}
      \hfill{\tiny Earnings-22, TN}\\[1pt]
      \textbf{GT:} ``Okay, 12\% plus, that is higher in around 50 basis points.''\\[0.5pt]
      \textbf{Instr:} {\itshape Normalize numbers and symbols to spoken form.}\\[0.5pt]
      \textbf{Pref:} ``Okay, \textbf{twelve percent} plus, that is higher in around \textbf{fifty} basis points.''}};

    \node[exbox, fill=entbg, draw=entbdr, below=2pt of norm] (ent) {\parbox{\exw}{%
      \textbf{\textcolor{entbdr}{Entities}}
      \hfill{\tiny Common Voice, with false positives}\\[1pt]
      \textbf{GT:} ``The event was hosted by Geograph's sponsor, Ordnance Survey.''\\[0.5pt]
      \textbf{Instr:} {\itshape Audio may contain: Geograph, Ordnance Survey,
        \underline{National Geographic}, \underline{British Geological Survey}.}\\[0.5pt]
      \textbf{Pref:} ``The event was hosted by Geograph's sponsor, Ordnance Survey.''}};

    \node[exbox, fill=disfbg, draw=disfbdr, below=2pt of ent] (disf) {\parbox{\exw}{%
      \textbf{\textcolor{disfbdr}{Disfluencies}}
      \hfill{\tiny AMI, remove}\\[1pt]
      \textbf{GT:} ``I should \textbf{uh}, \textbf{you know}, at least very least have \textbf{uh um} a GUI and \textbf{uh} wire I'll be wiring that in hopefully tomorrow morning.''\\[0.5pt]
      \textbf{Instr:} {\itshape Remove hesitation markers like uh, um, ah, and ignore repeated words if any.}\\[0.5pt]
      \textbf{Pref:} ``I should at least very least have a GUI and wire I'll be wiring that in hopefully tomorrow morning.''}};

    \node[exbox, fill=casebg, draw=casebdr, below=2pt of disf] (cas) {\parbox{\exw}{%
      \textbf{\textcolor{casebdr}{Case}}
      \hfill{\tiny GigaSpeech, lower case}\\[1pt]
      \textbf{GT:} ``the governor's race would be the culmination of his life's work.''\\[0.5pt]
      \textbf{Instr:} {\itshape Output the transcription in lowercase format.}\\[0.5pt]
      \textbf{Pref:} ``the governors race would be the culmination of his lifes work''}};

    \end{tikzpicture}
    \vspace{-4pt}
    \caption{Examples from Preference-ASR. Each box shows ground truth (GT), preference instruction, and expected output (Pref). Bold marks transformations; \underline{underlined} entities are false positives absent from audio.}
    \label{fig:examples}
\end{figure}

\subsection{Dataset Construction Pipeline}

We build Preference-ASR through a two-stage pipeline (Figure~\ref{fig:pipeline}) designed to maximize coverage across preference categories while maintaining accuracy. We start with approximately 600 samples from each of the seven source datasets listed in Table~\ref{tab:dataset_stats}, giving a pool of 3{,}545 samples whose baseline transcripts were manually verified and corrected by human annotators.

\textbf{Stage~1: Preference Classification.}
We use Qwen3-30B-A3B\footnote{\url{https://huggingface.co/Qwen/Qwen3-30B-A3B-Instruct-2507-FP8}}~\cite{qwen3technicalreport} for both stages. In stage~1, the model classifies each transcription by which preference categories it contains. For instance, a sample with ``three point five dollars'' is tagged under normalization (numbers and symbols), while one containing ``um'' or word repetitions is flagged for disfluencies.

\textbf{Stage~2: Instruction and Reference Generation.}
For each classified sample from stage~1, the model generates one or two distinct instructions along with the corresponding ``preference texts.'' We produce both directions where applicable: TN and ITN for normalization, keep or remove for disfluencies. For entities and casing, the preference text matches the ground-truth transcription and we generate instructions accordingly.

To keep each sample in only one category we apply a priority ordering: normalization first, then entities, disfluencies, and finally case. This reflects the relative complexity of the alignment challenge, with normalization and entities being the hardest to verify and case being the most straightforward.

\subsection{Dataset Statistics and Distribution}

After construction and deduplication, the dataset contains 3{,}210 unique audio-instruction-reference triples; the remaining 335 samples serve as standard baseline test cases without preference instructions. Table~\ref{tab:dataset_stats} shows the distribution across datasets and categories.

\begin{table}[t]
    \centering
    \caption{Distribution of unique preference triples across seven source datasets and four preference categories after priority-based deduplication including both directions. Original test sets are obtained from leaderboard\protect\footnotemark\ to keep evals reproducible.}
    \label{tab:dataset_stats}
    \footnotesize
    \begin{tabular}{@{}lrrrrr@{}}
        \toprule
        \textbf{Dataset} & \textbf{Norm.} & \textbf{Ent.} & \textbf{Disfl.} & \textbf{Case} & \textbf{Total} \\
        \midrule
        AMI~\cite{ami}          &  25 &   6 & 163 & 213 &  407 \\
        Common Voice~\cite{ardila2020commonvoice} &  17 &  98 &   3 & 234 &  352 \\
        Earnings-22~\cite{delrio2022earnings22}  & 102 &  37 & 108 &  90 &  337 \\
        GigaSpeech~\cite{chen2021gigaspeech}   &  88 & 118 & 132 & 181 &  519 \\
        LibriSpeech~\cite{panayotov2015librispeech}  &  23 &  92 &  34 & 445 &  594 \\
        SPGISpeech~\cite{oneill2021spgispeech}   & 105 &  72 &  41 & 198 &  416 \\
        VoxPopuli~\cite{wang2021voxpopuli}    &  49 & 257 &  58 & 221 &  585 \\
        \midrule
        \textbf{Total} & \textbf{409} & \textbf{680} & \textbf{539} & \textbf{1{,}582} & \textbf{3{,}210} \\
        \bottomrule
    \end{tabular}
\end{table}
\footnotetext{\url{https://huggingface.co/spaces/hf-audio/open_asr_leaderboard}}

The distribution reflects the nature of each source corpus. AMI~\cite{ami} and GigaSpeech~\cite{chen2021gigaspeech} contribute heavily to disfluencies because of their spontaneous, conversational style. VoxPopuli~\cite{wang2021voxpopuli} supplies the bulk of entity samples, consistent with its parliamentary domain. SPGISpeech~\cite{oneill2021spgispeech} and Earnings-22~\cite{delrio2022earnings22} dominate normalization, as expected from financial transcripts. This natural variation ensures the benchmark exercises a broad range of real-world transcription challenges.
\section{Experiment Setup}
\label{sec:experiments}

We evaluate the Preference-ASR dataset along two dimensions: (1)~how providing a preference instruction affects raw transcription accuracy under standard normalization, and (2)~whether models actually comply with the requested preference when assessed with a preference-aware metric.

\subsection{Preference-Aware Normalizer}

Standard WER pipelines~\cite{open_asr_leaderboard_normalizer_github_2026} apply blanket text normalization (lowercasing, converting numbers to words, stripping symbols) before computing edit distance~\cite{radford2023whisper}. This erases exactly the formatting distinctions that preference instructions target: if a model correctly outputs ``22nd'' following an ITN instruction, a standard normalizer converts it to ``twenty second,'' penalizing correct behavior~\cite{gupta2024normalization}.

We propose a \textit{preference-aware normalizer} that selectively disables the steps corresponding to the active preference. For \textit{normalization} preferences, it skips its own TN or ITN step so that spoken-form (e.g., ``twenty two'') or written-form (e.g., ``22nd'') outputs are compared as-is. For \textit{disfluency} preferences requesting retention, it preserves repetitions and fillers in both reference and hypothesis; when removal is requested, it strips them from both. We also normalize all filler words to ``um'' for fair comparison across models. 
For \textit{case} preferences, it suppresses lowercasing so that capitalization and punctuation are directly evaluated. For \textit{entities}, standard normalization is applied, since WER naturally captures entity-name accuracy. All non-conflicting steps (spelling standardization, whitespace cleanup) remain active regardless of preference type.

\subsection{Models}

We benchmark four models chosen to cover a range of instruction-following ability. \textit{Parakeet-TDT-0.6B-v3}~\cite{sekoyan2025canary} is a traditional non-LLM baseline: a FastConformer~\cite{rekesh2023fast} encoder with a TDT~\cite{xu2023efficient} decoder using greedy decoding and no instruction input. \textit{Canary-Qwen-2.5B}~\cite{sekoyan2025canary,qwen2024qwen2} pairs a FastConformer encoder with a Qwen~2.5 LLM backend via LoRA~\cite{hu2022lora} adaptation but was trained on standard ASR pairs without preference-specific tuning. The remaining two models natively support instructions: \textit{Phi-4-Multimodal}~\cite{abouelenin2025phi4} (5.6B), which integrates speech, vision, and text through modality-specific LoRA experts 
and \textit{Qwen3-Omni-30B}~\cite{xu2025qwen3omni}, a natively multimodal model trained with contextual biasing, representing the strongest instruction-aware system we evaluate. All inference and dataset generation were conducted on two 48GB NVIDIA A6000 GPUs.

\begin{table*}[!th]
    \centering
    \caption{WER~(\%) on Preference-ASR. For each model, \textbf{Std}\,=\,standard WER with full normalization; \textbf{Pref}\,=\,preference-aware WER with selective normalization that skips steps matching the active preference. Each preference category shows results under a \textit{default} prompt (no preference directive) and an \textit{instruction} prompt (preference instruction appended).}
    \label{tab:wer_results}
    \footnotesize
    \setlength{\tabcolsep}{1pt}
    \begin{tabular}{@{}c c cc cc cc cc cc cc@{}}
        \toprule
        \multirow{2}{*}{\textbf{Model}} & \multirow{2}{*}{\textbf{WER Eval}} & \multicolumn{2}{c}{\textbf{Normalization}} & \multicolumn{2}{c}{\textbf{Entities}} & \multicolumn{2}{c}{\textbf{Disfluencies}} & \multicolumn{2}{c}{\textbf{Case}} & \multicolumn{2}{c}{\textbf{Standard}} & \multicolumn{2}{c}{\textbf{Overall}} \\
        \cmidrule(lr){3-4} \cmidrule(lr){5-6} \cmidrule(lr){7-8} \cmidrule(lr){9-10} \cmidrule(lr){11-12} \cmidrule(lr){13-14}
        & & default & instruction & default & instruction & default & instruction & default & instruction & default & instruction & default & instruction \\
        \midrule
        \multirow{2}{*}{Parakeet-tdt-0.6b-v3~\cite{sekoyan2025canary}}
          & Std  &  6.24 & -- &  5.26 & -- & 10.65 & -- &  3.75 & -- & 4.81 & -- &  6.09 & -- \\
          & Pref & 11.16 & -- &  4.97 & -- & 10.93 & -- &  9.40 & -- & 4.77 & -- &  8.35 & -- \\
        \midrule
        \multirow{2}{*}{Canary-Qwen-2.5b~\cite{canaryqwen}}
          & Std  &  5.54 &  6.43 &  5.04 &  5.17 & 10.04 &  9.98 &  3.59 &  3.60 & 3.33 &  3.33 &  5.64 &  5.84 \\
          & Pref & 10.56 & 11.32 &  4.78 &  4.90 & 10.49 & 10.63 & 10.08 & 10.04 & 3.30 &  3.30 &  8.16 &  8.36 \\
        \midrule
        \multirow{2}{*}{Phi-4-multimodal~\cite{abouelenin2025phi4}}
          & Std  &  5.95 &  6.23 &  5.50 &  5.50 & 50.18 & 10.46 &  3.93 & 19.76 & 4.41 &  3.88 & 13.79 &  9.95 \\
          & Pref & 10.76 & 11.10 &  5.26 &  5.28 & 49.88 & 10.79 &  5.88 & 27.23 & 4.39 &  3.87 & 15.11 & 12.64 \\
        \midrule
        \multirow{2}{*}{Qwen3-Omni-30B~\cite{xu2025qwen3omni}}
          & Std  &  6.00 &  5.25 &  5.12 & 12.85 &  9.87 &  9.28 &  3.32 &  3.09 & 3.40 &  3.46 &  5.66 &  7.81 \\
          & Pref & 10.90 &  9.84 &  4.84 & 12.68 & 10.83 &  9.90 & 10.01 &  9.82 & 3.40 &  3.46 &  8.30 & 10.34 \\
        \bottomrule
    \end{tabular}
\end{table*}

\section{Evaluation Results}

Each model is evaluated under two settings: \textit{default}~(D), a standard ASR prompt without any preference instruction, and \textit{instructed}~(I), with a preference-specific instruction appended. Parakeet does not accept instruction input and is therefore evaluated only in the default setting. We report WER~(\%) grouped by preference category.

\textbf{Standard WER.}
Table~\ref{tab:wer_results} reports WER with full standard normalization applied to both reference and hypothesis, isolating how instructions affect raw word accuracy independent of formatting.

Under standard normalization, Canary-Qwen (5.64\%) and Qwen3-Omni (5.66\%) achieve nearly identical default WER, both outperforming Parakeet (6.09\%). Their responses to instructions, however, diverge sharply. Qwen3-Omni's entity WER spikes from 5.12\% to 12.85\% when instructed: the model inserts entity names from the prompt even when they are absent from the audio, a clear case of prompt-driven hallucination. On normalization (6.00\%\,$\to$\,5.25\%) and case (3.32\%\,$\to$\,3.09\%), the same model improves with explicit formatting cues. Canary-Qwen, by contrast, shows minimal sensitivity to instructions across all categories (5.64\%\,$\to$\,5.84\% overall), suggesting its LLM backend does not transfer instruction-following to the audio domain without dedicated alignment training.

Phi-4 exhibits a striking pattern on disfluencies: its default WER of 50.18\% indicates the model aggressively removes disfluencies by default, but when explicitly instructed it drops to 10.46\%, showing strong compliance for this category. However, case instructions backfire (3.93\%\,$\to$\,19.76\%), suggesting the model misinterprets casing directives. 

On the 335 non-preference baseline samples (``Standard'' column), the default and instructed settings use slightly different generic prompts (``Transcribe the English audio into text'' vs.\ ``Transcribe the speech in the input English audio'') with no preference directive. Even this minor rephrasing causes measurable WER shifts in some models (e.g., Phi-4: 4.41\%\,$\to$\,3.88\%), showing how sensitive SpeechLLMs can be to prompt wording. Canary-Qwen remains entirely invariant (3.33\%), confirming that preference prompts do not degrade standard recognition, as it was solely trained to perform ASR.

\textbf{Preference-Aware WER.}
The Pref rows of Table~\ref{tab:wer_results} report WER with our selective normalizer, directly measuring whether models comply with the requested formatting preference. 

Once we switch to the selective normalizer, quality differences that standard evaluation hides become visible. WER rises across all models; for example, Canary-Qwen's case WER jumps from 3.59\% to 10.08\% in the default setting, revealing that its output formatting deviates from the expected casing even when word content is largely correct.

Model rankings also shift when instructions are provided. In normalization, Qwen3-Omni\,(I) at 9.84\% clearly outperforms both Canary-Qwen\,(I) at 11.32\% and Parakeet at 11.16\%, a gap entirely invisible under standard normalization. For disfluencies, Qwen3-Omni\,(I) achieves 9.90\%, beating all default baselines including Parakeet (10.93\%) and showing genuine instruction compliance. Phi-4's disfluency improvement carries over here as well (49.88\%\,$\to$\,10.79\%), but its case WER worsens sharply (5.88\%\,$\to$\,27.23\%), confirming that its instruction-following is category-dependent rather than general.

The entity hallucination problem persists under preference-aware evaluation (12.68\%), confirming it is not a normalization artifact but a real failure to ground predictions in the audio. This highlights a key open challenge: current SpeechLLMs struggle to balance textual prompt context against acoustic evidence, which is critical for context-biased ASR deployment.

Taken together, these results show that Preference-ASR surfaces tradeoffs invisible to traditional benchmarks. Qwen3-Omni excels at formatting compliance for normalization, disfluencies, and case, but its entity handling degrades when instruction context conflicts with the audio. Canary-Qwen's flat response to instructions confirms that an LLM backbone alone, without preference-specific training, is not enough for instruction-following ASR.

\section{Conclusion}

We introduced Preference-ASR, a test set of 3{,}210 samples drawn from seven open-source corpora that evaluates whether ASR systems can follow explicit human preference instructions across four categories: normalization, entities, disfluencies, and case. Together with a preference-aware normalizer that selectively skips steps matching the active instruction, the benchmark exposes quality differences that standard WER evaluation hides. Our experiments show that model rankings shift substantially depending on the preference type. Qwen3-Omni follows formatting instructions well for normalization, disfluencies, and case, but hallucinates entity names when biasing terms appear in the prompt, a failure mode that traditional benchmarks cannot detect. Canary-Qwen's flat response to instructions shows that an LLM backbone alone, without preference-aligned training, is not enough.

Several limitations remain: the dataset covers only English, does not yet include multi-speaker preferences, and the LLM-generated preference texts required manual verification, especially for normalization where acoustic context cannot be inferred from text alone. As SpeechLLMs grow more capable of following instructions, benchmarks must evolve to test that capability directly. 
In future work we plan to reduce the manual effort in the pipeline, add single and multi-speaker preference instructions, and extend the benchmark to multilingual settings.

\section{Generative AI Use Disclosure}
Generative AI tools (LLMs) were used in two capacities: (1)~as part of the dataset construction pipeline for preference classification and instruction generation, and (2)~for editing and polishing the manuscript. All content was reviewed and validated by the authors.

\bibliographystyle{IEEEtran}
\bibliography{mybib}

\onecolumn

\begin{center}
  {\large\textbf{Supplementary Material: Preference-ASR}}\\[0.3em]
\end{center}
\vspace{0.5em}

\section{Additional Preference Examples}

Figure~3 in the main paper shows one example per preference category. Figure~\ref{fig:supp_examples} lists a sample from each category along with the full instruction and expected preference text as used during evaluation.

\begin{center}
\begin{tikzpicture}[
  font=\scriptsize,
  exbox/.style={draw, rounded corners=2pt, semithick, inner sep=3pt},
]
\node[exbox, fill=normbg, draw=normbdr] (norm) {\parbox{0.93\textwidth}{%
  \textbf{\textcolor{normbdr}{Normalization}}
  \hfill{\tiny AMI, ITN, numbers \& symbols}\\[1pt]
  \textbf{GT:} ``Um, and profit aim is fifty million Euros, which is uh''\\[0.5pt]
  \textbf{Instr:} {\itshape Perform speech recognition for provided audio. Output the transcription. Inverse Normalize numbers and symbols to written form. Example: In written form, use numerical digits and symbols as is instead of in words like 22 for twenty two and 15 for fifteen.}\\[0.5pt]
  \textbf{Pref:} ``Um, and profit aim is \textbf{\euro\,50,000,000} which is uh''}};
\node[exbox, fill=entbg, draw=entbdr, below=2pt of norm] (ent) {\parbox{0.93\textwidth}{%
  \textbf{\textcolor{entbdr}{Entities}}
  \hfill{\tiny Earnings-22, with false positives}\\[1pt]
  \textbf{GT:} ``I'll now hand the call over to Craig Larson, Head of Investor Relations for KKR.''\\[0.5pt]
  \textbf{Instr:} {\itshape Transcribe the audio. Audio may contain entities: Craig Larson, KKR, \underline{Michael Smith}, \underline{Blackstone}, \underline{Goldman Sachs}.}\\[0.5pt]
  \textbf{Pref:} ``I'll now hand the call over to Craig Larson, Head of Investor Relations for KKR.''}};
\node[exbox, fill=disfbg, draw=disfbdr, below=2pt of ent] (disf) {\parbox{0.93\textwidth}{%
  \textbf{\textcolor{disfbdr}{Disfluencies}}
  \hfill{\tiny AMI, keep}\\[1pt]
  \textbf{GT:} ``Actually is the code accessible, \textbf{like} \textbf{the the} GUI stuff that you've done, \textbf{yeah}.''\\[0.5pt]
  \textbf{Instr:} {\itshape Transcribe the input speech. Keep all words spoken by the speaker as is including filler words and repeated words.}\\[0.5pt]
  \textbf{Pref:} ``Actually is the code accessible, \textbf{like} \textbf{the the} GUI stuff that you've done, \textbf{yeah}.''}};
\node[exbox, fill=casebg, draw=casebdr, below=2pt of disf] (cas) {\parbox{0.93\textwidth}{%
  \textbf{\textcolor{casebdr}{Case}}
  \hfill{\tiny Common Voice, proper punctuated \& capitalized}\\[1pt]
  \textbf{GT:} ``These options now provide potentially useful drugs for treating this iron overload problem.''\\[0.5pt]
  \textbf{Instr:} {\itshape Transcribe the speech in the input audio. Output with correct casing and punctuation.}\\[0.5pt]
  \textbf{Pref:} ``\textbf{T}hese options now provide potentially useful drugs for treating this iron overload problem\textbf{.}''}};
\end{tikzpicture}
\captionof{figure}{Additional examples from Preference-ASR (complementary to Figure~3 in the main paper). Normalization shows the ITN direction (written form); Disfluencies shows the \emph{keep} direction (preserve verbatim speech). Bold marks key elements; \underline{underlined} entities are false positives absent from audio.}
\label{fig:supp_examples}
\end{center}

\section{Per-Dataset WER Breakdown}

Table~2 in the main paper reports aggregate WER across all seven source datasets. Tables~\ref{tab:supp_canary}--\ref{tab:supp_qwen3} provide the full per-dataset breakdown for each model, reporting both standard WER (\textbf{Std}, full normalization) and preference-aware WER (\textbf{Pref}, selective normalization that skips steps matching the active preference). D denotes a default prompt with no preference directive; I denotes a prompt with the preference instruction appended. Parakeet does not accept instructions and is evaluated only under the default setting.


\vspace{0.5em}

\begin{center}
    \footnotesize
    \setlength{\tabcolsep}{0.5pt}
    \captionof{table}{Per-dataset WER~(\%) for \textbf{Canary-Qwen-2.5b}. \textbf{Std}\,=\,standard WER; \textbf{Pref}\,=\,preference-aware WER. D\,=\,default prompt; I\,=\,instruction prompt.}
    \label{tab:supp_canary}
    \vspace{0.3em}
    \begin{tabular*}{\textwidth}{@{\extracolsep{\fill}}l c cc cc cc cc cc cc@{}}
        \toprule
        \multirow{2}{*}{\textbf{Dataset}} & \multirow{2}{*}{\textbf{WER}} & \multicolumn{2}{c}{\textbf{Norm.}} & \multicolumn{2}{c}{\textbf{Entities}} & \multicolumn{2}{c}{\textbf{Disfl.}} & \multicolumn{2}{c}{\textbf{Case}} & \multicolumn{2}{c}{\textbf{Std.}} & \multicolumn{2}{c}{\textbf{Overall}} \\
        \cmidrule(lr){3-4} \cmidrule(lr){5-6} \cmidrule(lr){7-8} \cmidrule(lr){9-10} \cmidrule(lr){11-12} \cmidrule(lr){13-14}
        & & D & I & D & I & D & I & D & I & D & I & D & I \\
        \midrule
        \multirow{2}{*}{AMI}
          & Std  &  6.24 & 23.23 &  5.81 &  7.56 &  9.75 & 10.00 &  8.11 &  7.62 &  8.70 &  8.70 &  9.04 & 10.53 \\
          & Pref & 14.29 & 29.59 &  5.81 &  7.56 &  9.13 &  9.39 & 32.20 & 30.81 &  8.70 &  8.70 & 13.33 & 14.60 \\
        \midrule
        \multirow{2}{*}{CommonVoice}
          & Std  &  6.58 &  7.12 &  8.17 &  6.80 & 13.85 & 13.85 &  4.46 &  4.37 & 11.48 & 11.48 &  6.43 &  5.84 \\
          & Pref & 12.16 & 12.43 &  8.17 &  6.80 & 13.85 & 13.85 &  7.39 &  7.65 & 11.48 & 11.48 &  8.23 &  7.78 \\
        \midrule
        \multirow{2}{*}{Earnings-22}
          & Std  &  7.52 &  8.33 &  8.84 &  8.41 &  8.53 &  8.89 &  4.51 &  4.60 & 10.13 & 11.39 &  7.81 &  8.24 \\
          & Pref & 14.04 & 14.73 &  8.73 &  8.30 & 11.18 & 12.08 & 14.38 & 11.80 & 10.13 & 11.39 & 12.34 & 12.68 \\
        \midrule
        \multirow{2}{*}{GigaSpeech}
          & Std  &  9.24 & 10.71 &  8.58 &  9.48 & 15.83 & 15.27 &  8.65 &  9.07 &  7.69 &  7.24 & 10.72 & 11.17 \\
          & Pref & 14.02 & 15.07 &  7.19 &  8.07 & 15.28 & 15.00 & 16.14 & 17.19 &  7.52 &  7.07 & 12.38 & 12.92 \\
        \midrule
        \multirow{2}{*}{LibriSpeech}
          & Std  &  2.77 &  2.49 &  2.54 &  2.62 &  8.80 &  8.71 &  1.80 &  1.83 &  3.73 &  4.35 &  2.68 &  2.70 \\
          & Pref &  7.62 &  7.35 &  2.55 &  2.62 &  8.56 &  8.48 & 10.88 & 10.81 &  3.73 &  4.35 &  8.11 &  8.07 \\
        \midrule
        \multirow{2}{*}{SPGISpeech}
          & Std  &  2.14 &  2.05 &  2.18 &  2.05 &  4.09 &  4.18 &  1.18 &  1.14 &  1.68 &  1.72 &  2.03 &  2.00 \\
          & Pref &  5.98 &  5.95 &  2.18 &  2.05 &  4.09 &  4.18 &  3.63 &  3.53 &  1.68 &  1.72 &  3.63 &  3.59 \\
        \midrule
        \multirow{2}{*}{VoxPopuli}
          & Std  &  3.93 &  3.89 &  4.25 &  4.40 &  6.66 &  6.44 &  3.87 &  3.78 &  2.99 &  2.99 &  4.39 &  4.43 \\
          & Pref &  8.23 &  8.20 &  4.23 &  4.38 &  7.92 &  7.70 &  6.90 &  6.94 &  2.99 &  2.99 &  5.58 &  5.65 \\
        \midrule
        \multirow{2}{*}{\textbf{Overall}}
          & Std  &  5.54 &  6.43 &  5.04 &  5.17 & 10.04 &  9.98 &  3.59 &  3.60 &  3.33 &  3.33 &  5.64 &  5.84 \\
          & Pref & 10.56 & 11.32 &  4.78 &  4.90 & 10.49 & 10.63 & 10.08 & 10.04 &  3.30 &  3.30 &  8.16 &  8.36 \\
        \bottomrule
    \end{tabular*}
\end{center}

\vspace{1em}

\begin{center}
    \footnotesize
    \setlength{\tabcolsep}{0.5pt}
    \captionof{table}{Per-dataset WER~(\%) for \textbf{Parakeet-tdt-0.6b-v3}. This model does not accept instructions, so only default-prompt results are shown.}
    \label{tab:supp_parakeet}
    \vspace{0.3em}
    \begin{tabular*}{\textwidth}{@{\extracolsep{\fill}}l c cc cc cc cc cc cc@{}}
        \toprule
        \multirow{2}{*}{\textbf{Dataset}} & \multirow{2}{*}{\textbf{WER}} & \multicolumn{2}{c}{\textbf{Norm.}} & \multicolumn{2}{c}{\textbf{Entities}} & \multicolumn{2}{c}{\textbf{Disfl.}} & \multicolumn{2}{c}{\textbf{Case}} & \multicolumn{2}{c}{\textbf{Std.}} & \multicolumn{2}{c}{\textbf{Overall}} \\
        \cmidrule(lr){3-4} \cmidrule(lr){5-6} \cmidrule(lr){7-8} \cmidrule(lr){9-10} \cmidrule(lr){11-12} \cmidrule(lr){13-14}
        & & D & I & D & I & D & I & D & I & D & I & D & I \\
        \midrule
        \multirow{2}{*}{AMI}
          & Std  &  7.10 & -- &  4.65 & -- & 10.35 & -- &  9.79 & -- & 14.67 & -- &  9.96 & -- \\
          & Pref & 14.69 & -- &  4.65 & -- & 10.15 & -- & 16.45 & -- & 14.67 & -- & 11.57 & -- \\
        \midrule
        \multirow{2}{*}{CommonVoice}
          & Std  &  6.03 & -- &  9.89 & -- & 23.08 & -- &  5.68 & -- & 16.39 & -- &  7.89 & -- \\
          & Pref & 11.62 & -- &  9.89 & -- & 23.08 & -- &  8.05 & -- & 16.39 & -- &  9.42 & -- \\
        \midrule
        \multirow{2}{*}{Earnings-22}
          & Std  &  8.19 & -- &  8.72 & -- &  9.94 & -- &  5.34 & -- & 12.66 & -- &  8.67 & -- \\
          & Pref & 14.43 & -- &  8.36 & -- & 10.68 & -- & 10.78 & -- & 12.66 & -- & 12.01 & -- \\
        \midrule
        \multirow{2}{*}{GigaSpeech}
          & Std  &  9.76 & -- &  8.58 & -- & 16.55 & -- &  8.83 & -- &  4.95 & -- & 10.90 & -- \\
          & Pref & 14.23 & -- &  7.19 & -- & 15.86 & -- & 17.52 & -- &  4.77 & -- & 12.66 & -- \\
        \midrule
        \multirow{2}{*}{LibriSpeech}
          & Std  &  3.60 & -- &  3.91 & -- &  8.80 & -- &  2.04 & -- &  4.35 & -- &  3.25 & -- \\
          & Pref &  8.44 & -- &  3.87 & -- &  8.65 & -- & 10.88 & -- &  4.35 & -- &  8.53 & -- \\
        \midrule
        \multirow{2}{*}{SPGISpeech}
          & Std  &  3.93 & -- &  4.55 & -- &  6.54 & -- &  2.08 & -- &  4.23 & -- &  3.96 & -- \\
          & Pref &  7.82 & -- &  4.45 & -- &  8.22 & -- &  4.96 & -- &  4.20 & -- &  5.82 & -- \\
        \midrule
        \multirow{2}{*}{VoxPopuli}
          & Std  &  3.01 & -- &  3.58 & -- &  4.79 & -- &  2.32 & -- &  2.40 & -- &  3.40 & -- \\
          & Pref &  7.56 & -- &  3.57 & -- &  6.11 & -- &  4.97 & -- &  2.40 & -- &  4.55 & -- \\
        \midrule
        \multirow{2}{*}{\textbf{Overall}}
          & Std  &  6.24 & -- &  5.26 & -- & 10.65 & -- &  3.75 & -- &  4.81 & -- &  6.09 & -- \\
          & Pref & 11.16 & -- &  4.97 & -- & 10.93 & -- &  9.40 & -- &  4.77 & -- &  8.35 & -- \\
        \bottomrule
    \end{tabular*}
\end{center}

\vspace{1em}

\begin{center}
    \footnotesize
    \setlength{\tabcolsep}{0.5pt}
    \captionof{table}{Per-dataset WER~(\%) for \textbf{Phi-4-multimodal}. $^\dagger$Extreme WER values caused by the model generating extraneous non-transcription text for certain Earnings-22 samples.}
    \label{tab:supp_phi4}
    \vspace{0.3em}
    \begin{tabular*}{\textwidth}{@{\extracolsep{\fill}}l c cc cc cc cc cc cc@{}}
        \toprule
        \multirow{2}{*}{\textbf{Dataset}} & \multirow{2}{*}{\textbf{WER}} & \multicolumn{2}{c}{\textbf{Norm.}} & \multicolumn{2}{c}{\textbf{Entities}} & \multicolumn{2}{c}{\textbf{Disfl.}} & \multicolumn{2}{c}{\textbf{Case}} & \multicolumn{2}{c}{\textbf{Std.}} & \multicolumn{2}{c}{\textbf{Overall}} \\
        \cmidrule(lr){3-4} \cmidrule(lr){5-6} \cmidrule(lr){7-8} \cmidrule(lr){9-10} \cmidrule(lr){11-12} \cmidrule(lr){13-14}
        & & D & I & D & I & D & I & D & I & D & I & D & I \\
        \midrule
        \multirow{2}{*}{AMI}
          & Std  &  7.96 &  5.81 & 12.79 & 10.47 & 10.30 & 10.15 & 10.39 & 10.68 & 16.85 & 19.57 & 10.41 & 10.20 \\
          & Pref & 14.29 & 13.27 & 12.79 & 10.47 & 10.32 & 10.15 & 17.75 & 18.15 & 16.85 & 19.02 & 12.17 & 12.04 \\
        \midrule
        \multirow{2}{*}{CommonVoice}
          & Std  &  6.58 &  8.22 &  9.60 &  5.61 & 20.00 & 16.92 &  4.50 &  4.77 & 14.75 & 16.39 &  7.19 &  5.69 \\
          & Pref & 12.16 & 13.78 &  9.60 &  5.61 & 13.85 & 13.85 &  6.86 &  7.48 & 14.75 & 16.39 &  8.64 &  7.35 \\
        \midrule
        \multirow{2}{*}{Earnings-22}
          & Std  &  7.33 &  7.88 &  6.78 &  5.93 & 201.55$^\dagger$ &  9.90 &  6.44 & 382.80$^\dagger$ & 13.92 & 14.56 & 74.15 & 41.39 \\
          & Pref & 13.69 & 14.54 &  6.67 &  5.82 & 195.18$^\dagger$ & 10.95 & 10.69 & 515.21$^\dagger$ & 13.92 & 14.56 & 75.63 & 54.90 \\
        \midrule
        \multirow{2}{*}{GigaSpeech}
          & Std  &  9.07 & 10.08 &  8.75 & 10.54 & 15.99 & 15.78 &  7.93 &  7.63 &  4.77 &  5.57 & 10.50 & 11.16 \\
          & Pref & 13.90 & 14.83 &  7.57 &  9.47 & 15.65 & 15.47 &  8.64 &  7.44 &  4.69 &  5.57 & 11.17 & 11.72 \\
        \midrule
        \multirow{2}{*}{LibriSpeech}
          & Std  &  2.49 &  2.21 &  3.96 &  2.84 &  9.21 &  9.63 &  2.56 &  2.68 &  4.35 &  4.97 &  3.54 &  3.32 \\
          & Pref &  7.35 &  7.07 &  3.92 &  2.80 &  8.98 &  9.39 &  2.55 &  2.70 &  4.35 &  4.97 &  3.75 &  3.55 \\
        \midrule
        \multirow{2}{*}{SPGISpeech}
          & Std  &  3.18 &  3.22 &  3.19 &  3.82 &  5.72 &  5.00 &  1.64 &  1.51 &  3.55 &  2.37 &  3.18 &  2.95 \\
          & Pref &  6.74 &  6.77 &  3.09 &  3.80 &  6.27 &  5.22 &  4.58 &  5.39 &  3.55 &  2.37 &  4.84 &  4.81 \\
        \midrule
        \multirow{2}{*}{VoxPopuli}
          & Std  &  4.97 &  4.65 &  4.50 &  4.59 &  6.16 &  6.13 &  3.55 &  3.42 &  2.40 &  2.59 &  4.52 &  4.52 \\
          & Pref &  8.97 &  8.39 &  4.48 &  4.58 &  7.49 &  7.46 &  6.73 &  6.66 &  2.40 &  2.59 &  5.71 &  5.69 \\
        \midrule
        \multirow{2}{*}{\textbf{Overall}}
          & Std  &  5.95 &  6.23 &  5.50 &  5.50 & 50.18$^\dagger$ & 10.46 &  3.93 & 19.76$^\dagger$ &  4.41 &  3.88 & 13.79 &  9.95 \\
          & Pref & 10.76 & 11.10 &  5.26 &  5.28 & 49.88$^\dagger$ & 10.79 &  5.88 & 27.23$^\dagger$ &  4.39 &  3.87 & 15.11 & 12.64 \\
        \bottomrule
    \end{tabular*}
\end{center}

\vspace{1em}

\begin{center}
    \footnotesize
    \setlength{\tabcolsep}{0.5pt}
    \captionof{table}{Per-dataset WER~(\%) for \textbf{Qwen3-Omni-30B}. D\,=\,default prompt; I\,=\,instruction prompt.}
    \label{tab:supp_qwen3}
    \vspace{0.3em}
    \begin{tabular*}{\textwidth}{@{\extracolsep{\fill}}l c cc cc cc cc cc cc@{}}
        \toprule
        \multirow{2}{*}{\textbf{Dataset}} & \multirow{2}{*}{\textbf{WER}} & \multicolumn{2}{c}{\textbf{Norm.}} & \multicolumn{2}{c}{\textbf{Entities}} & \multicolumn{2}{c}{\textbf{Disfl.}} & \multicolumn{2}{c}{\textbf{Case}} & \multicolumn{2}{c}{\textbf{Std.}} & \multicolumn{2}{c}{\textbf{Overall}} \\
        \cmidrule(lr){3-4} \cmidrule(lr){5-6} \cmidrule(lr){7-8} \cmidrule(lr){9-10} \cmidrule(lr){11-12} \cmidrule(lr){13-14}
        & & D & I & D & I & D & I & D & I & D & I & D & I \\
        \midrule
        \multirow{2}{*}{AMI}
          & Std  &  6.88 &  6.24 & 10.47 &  6.40 & 10.25 &  9.98 &  9.99 & 10.29 & 11.96 & 12.50 & 10.00 &  9.71 \\
          & Pref & 15.10 & 14.49 & 10.47 &  6.40 & 10.81 &  9.71 & 18.64 & 19.34 & 11.96 & 12.50 & 12.51 & 11.72 \\
        \midrule
        \multirow{2}{*}{CommonVoice}
          & Std  &  7.12 &  6.03 &  6.37 &  2.85 & 10.77 & 10.77 &  4.55 &  4.46 & 22.95 & 22.95 &  5.84 &  4.20 \\
          & Pref & 12.70 & 11.62 &  6.37 &  2.85 & 10.77 & 10.77 &  6.90 &  7.08 & 22.95 & 22.95 &  7.37 &  5.85 \\
        \midrule
        \multirow{2}{*}{Earnings-22}
          & Std  &  7.86 &  6.94 &  8.78 &  8.66 &  8.58 &  7.73 &  3.96 &  3.86 & 11.39 & 11.39 &  7.93 &  7.23 \\
          & Pref & 14.43 & 13.08 &  8.67 &  8.55 & 11.86 & 10.61 & 10.05 & 10.05 & 11.39 & 11.39 & 12.38 & 11.35 \\
        \midrule
        \multirow{2}{*}{GigaSpeech}
          & Std  &  9.12 &  8.17 &  8.58 & 18.00 & 14.77 & 14.30 &  7.45 &  7.63 &  4.24 &  4.24 & 10.00 & 12.47 \\
          & Pref & 13.61 & 12.96 &  7.30 & 17.09 & 14.70 & 13.89 & 16.68 & 16.56 &  4.24 &  4.24 & 12.07 & 14.52 \\
        \midrule
        \multirow{2}{*}{LibriSpeech}
          & Std  &  2.49 &  2.77 &  2.74 & 13.10 &  8.55 &  8.46 &  1.47 &  1.29 &  2.48 &  2.48 &  2.49 &  5.30 \\
          & Pref &  7.35 &  7.62 &  2.75 & 13.14 &  8.31 &  8.23 & 10.51 & 10.16 &  2.48 &  2.48 &  7.90 & 10.61 \\
        \midrule
        \multirow{2}{*}{SPGISpeech}
          & Std  &  3.13 &  3.29 &  2.96 &  9.79 &  5.04 &  5.72 &  1.26 &  1.35 &  2.50 &  2.55 &  2.74 &  4.17 \\
          & Pref &  6.81 &  6.72 &  2.86 &  9.79 &  5.04 &  5.86 &  5.13 &  5.03 &  2.50 &  2.55 &  4.59 &  5.94 \\
        \midrule
        \multirow{2}{*}{VoxPopuli}
          & Std  &  4.55 &  2.45 &  4.36 & 13.46 &  6.23 &  4.36 &  3.47 &  2.42 &  1.80 &  2.00 &  4.38 &  8.93 \\
          & Pref &  8.62 &  5.38 &  4.30 & 13.46 &  7.55 &  5.58 &  9.09 &  8.70 &  1.80 &  2.00 &  6.01 & 10.56 \\
        \midrule
        \multirow{2}{*}{\textbf{Overall}}
          & Std  &  6.00 &  5.25 &  5.12 & 12.85 &  9.87 &  9.28 &  3.32 &  3.09 &  3.40 &  3.46 &  5.66 &  7.81 \\
          & Pref & 10.90 &  9.84 &  4.84 & 12.68 & 10.83 &  9.90 & 10.01 &  9.82 &  3.40 &  3.46 &  8.30 & 10.34 \\
        \bottomrule
    \end{tabular*}
\end{center}

\end{document}